\documentclass[letterpaper,10pt,conference]{ieeeconf}

\usepackage[T1]{fontenc}
\usepackage{times}
\usepackage{graphicx}
\usepackage{amsmath,amssymb}
\usepackage{bm}
\usepackage[space]{cite}
\usepackage{array}
\usepackage{booktabs}
\usepackage[skip=3pt,font={small}]{caption}
\usepackage[linkcolor=black,citecolor=black,urlcolor=black,colorlinks=true]{hyperref}
\DeclareGraphicsExtensions{.png,.jpg,.jpeg,.pdf}
\IEEEoverridecommandlockouts
\title{\LARGE \bf
Whole-Body Aerial Grasping and Lifting via Partial \\ Visual Observations}

\author{Jiaye Jin$^{*}$, Rui Jin$^{*}$, Xinhang Xu, Haotian Jin, Ruiyang Liu, Yi Wang, Jiayan Zhao, Kun Cao, and Lihua Xie$^\dagger$
    \thanks{$^*$Equal contribution.}
    \thanks{$^\dagger$Corresponding author: {\tt\small elhxie@ntu.edu.sg}.}
    \thanks{
    Jiaye Jin, Rui Jin, Xinhang Xu, Haotian Jin, and Ruiyang Liu are with the School of Electrical and Electronic Engineering, Nanyang Technological University, Singapore 639798.
    Yi Wang, Jiayan Zhao, and Kun Cao are with the College of Electronics and Information Engineering, Shanghai Institute of Intelligent Science and Technology, Tongji University, Shanghai 201804, China.
    Lihua Xie is with the NTU--VinUni Joint Research Laboratory for Embodied AI and Robotics, School of Electrical and Electronic Engineering, Nanyang Technological University, Singapore 639798, and VinUniversity, Hanoi, Vietnam.
    }
}

\begin{document}
\maketitle
\thispagestyle{empty}
\pagestyle{empty}
\suppressfloats[t]
\begin{abstract}
Aerial grasp-and-lift tasks require whole-body coordination across approach, acquisition, and lifting under partial target observations. Early approach failures can limit exposure to later task stages during training, while changing visibility complicates alignment and closure timing during execution. We present a recurrent teacher--student framework that learns a single policy in simulation to jointly command flight, arm motion, and gripper closure without an explicit task-phase input. A privileged teacher learns through reinforcement learning with a critical-state curriculum that exposes acquisition and lifting states before connecting them to normal approach trajectories. Its behavior is distilled into a recurrent visual student that replaces privileged target states with dual-view point clouds and proprioception, integrating observation history for closed-loop control. A dedicated closure objective supervises closure timing from sustained model-defined readiness sequences. Training and primary evaluation use a simulated acquisition-and-payload model with condition-triggered latching, virtual attachment, and wrench-based payload loading for short-distance lifting. Across 8,996 completed simulation episodes under this model, the frozen student achieves full-task success rates of 99.97\%, 97.14\%, and 95.84\% under nominal, physics/control-randomized, and additional camera-randomized conditions, respectively. The nominal latch-count-weighted mean of per-seed 90th-percentile alignment errors at acquisition is 8.12 mm.
\end{abstract}

\section{Introduction}
Uncrewed aerial manipulators (UAMs) extend UAV capabilities from observation~\cite{precisemaneuvers, yao2026arsgaussian,gsplanner} to physical interactions~\cite{floatdrone,tethered,handlike,ductam}.
Physical demonstrations show that onboard perception can support object retrieval~\cite{softvision}. Beyond reaching a suitable end-effector pose, the controller must coordinate relative motion, closure timing, and loaded flight. We study whole-body coordination and policy learning across approach, acquisition, and lifting using an environment-defined attachment and payload-wrench model.

\begin{figure}[t]
\centering
\IfFileExists{1.png}{%
\includegraphics[width=\columnwidth]{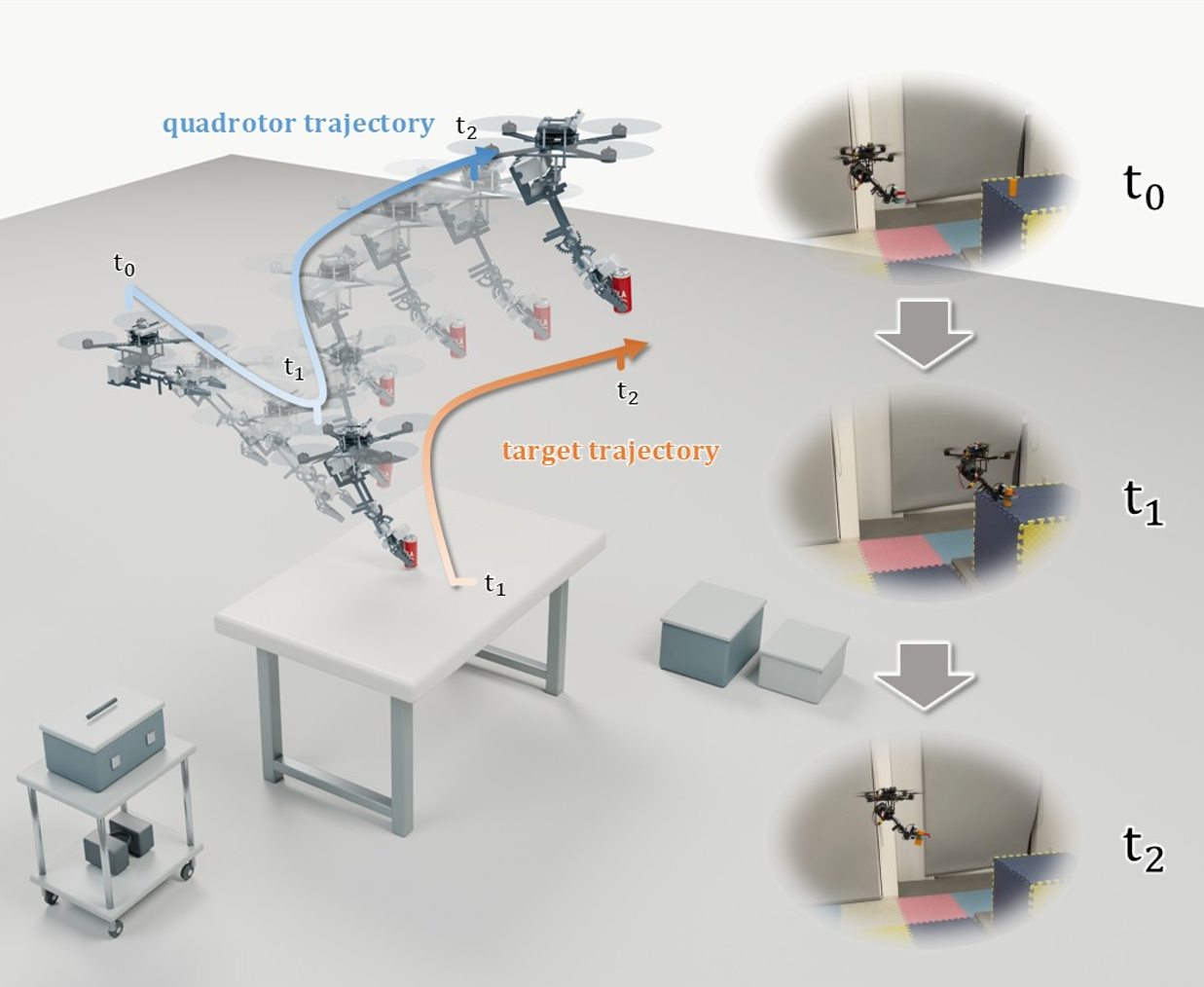}%
}{%
\fbox{\begin{minipage}[c][12mm][c]{\dimexpr\columnwidth-2\fboxsep-2\fboxrule\relax}
\centering\small\textbf{Complete task sequence --- reserved}
\end{minipage}}}
\caption{Aerial acquisition-and-lift scenario.
Left: simulated approach, modeled acquisition, and lifting.
Right: photographs of the physical platform and task scenario,
shown for illustration only.}
\label{fig:task_sequence}
\vspace{-0.5cm}
\end{figure}

The first challenge is coordinating the aerial base, arm, and gripper during acquisition and lifting. End-effector pose depends on vehicle motion and arm configuration; acquisition also constrains relative velocity, posture, and closure timing. Whole-body planning can enforce geometric and dynamic constraints \cite{wholebody}, but the feedback policy must coordinate motion and closure as observations change. Improving alignment can alter relative motion or disturb the base. After acquisition, the controller must stabilize the vehicle while lifting the added load.

The second challenge is discovering and connecting behaviors across the task. Early approach failures can prevent reinforcement learning from reaching successful closure or loaded flight, even with dense approach rewards. Goal relabeling and demonstration-guided exploration improve sparse-reward manipulation learning \cite{her,dapg}; this aerial task also requires reaching and connecting critical interaction states. Near-acquisition and post-acquisition starts expose later behaviors, but normal approach trajectories may not reach them. Learning must therefore address both skill discovery and skill connection, turning experience at critical states into successful trajectories from normal task starts.

The third challenge is maintaining closed-loop control under partial target observations. Visual feedback enables repeated grasp corrections \cite{handeye}, but vehicle and arm motion change body- and wrist-camera viewpoints and occlusions. A single point cloud may omit information needed to assess alignment or time closure. A recurrent policy can integrate visible geometry, proprioception, and observation history to retain context for motion and closure decisions as visibility changes during acquisition and lifting.

We develop a recurrent teacher--student framework for whole-body acquisition and lifting without an explicit task-phase input. A privileged teacher learns through reinforcement learning with near-acquisition, bridge, and post-acquisition resets, then connects these behaviors to normal approach starts. Its policy jointly commands the aerial base, three-DOF arm, and gripper. We distill this behavior into a visual student using dual-view point clouds and proprioception. A shared geometric encoder and recurrent state integrate target geometry and observation history. Short-horizon labels from sustained, model-defined acquisition readiness supervise closure, while the student retains the teacher's whole-body action interface.

Across 8,996 completed episodes under the acquisition-and-payload model, the frozen student achieves 99.97\%, 97.14\%, and 95.84\% full-task success under nominal, physics/control-randomized, and additional camera-randomized conditions. The nominal weighted mean of per-seed 90th-percentile alignment errors at acquisition is 8.12 mm. Nominal full-task success is close to the privileged teacher's under this model. Separate MuJoCo sim-to-sim trials assess three additional objects with virtual attachment or native contact. Physical-platform validation remains outside this study.

Our main contributions are:
\begin{itemize}
    \item A unified whole-body acquisition-and-lifting policy that jointly commands the aerial base, arm, and gripper without an explicit task-phase input.
    \item A critical-state curriculum for skill discovery through near-acquisition, bridge, and post-acquisition resets, followed by skill connection from normal task starts.
    \item Recurrent privileged-to-visual transfer with dual-view observations, memory, and closure supervision that retains the teacher's whole-body action interface.
\end{itemize}

\section{Related Works}
\subsection{Whole-Body Coordination for Aerial Manipulation}
Aerial manipulation uses aircraft-mounted grippers and articulated manipulators, whose mechanical coupling and interaction modes shape control requirements \cite{aerialreview}.

Compliant grippers accommodate positioning uncertainty: Fishman \emph{et al.} combine soft-gripper modeling with flight control, while Ubellacker \emph{et al.} extend this approach to onboard perception \cite{fishman,softvision}. With articulated arms, Deng \emph{et al.} coordinate base--arm trajectories through optimization subject to feasibility and collision constraints \cite{wholebody}.

Swooper incorporates gripper actuation into a learned flight policy for aerial grasping \cite{swooper}. Flying Hand executes teleoperated or learned end-effector commands through whole-body model predictive control \cite{flyinghand}. Our actor directly commands flight, arm joints, and the gripper from partial geometry under the simulated interaction model. Differences in interaction models and evaluation platforms preclude claims of superior physical grasping performance.

\subsection{Learning Manipulation with Geometric Feedback}
Hindsight Experience Replay reuses unsuccessful goal-conditioned trajectories by relabeling their goals \cite{her}, while Rajeswaran \emph{et al.} combine reinforcement learning and demonstrations for sample-efficient dexterous manipulation \cite{dapg}. Reverse curriculum generation expands training starts outward from a known goal to make sparse-success tasks accessible \cite{reverse}. SkiLD learns reusable skills and guides their composition with demonstrations \cite{skild}. Our curriculum exposes critical acquisition and lifting states, then connects these behaviors to normal approach starts.

Visual feedback introduces a training-to-deployment gap. Levine \emph{et al.} learn closed-loop hand--eye grasp coordination directly from camera images \cite{handeye}. Asymmetric Actor Critic exploits simulation state through a full-state critic and an image-based actor \cite{asymmetric}. UniDexGrasp distills privileged grasping policies into point-cloud-conditioned policies \cite{unidex}. Our student retains the teacher's action interface while replacing privileged target geometry with dual-view observations and recurrent state.

Temporal context can recover information absent from instantaneous observations. RMA infers an adaptation representation from recent proprioceptive and action histories for legged control \cite{rma}. It adapts to environmental dynamics; our recurrent state supports motion and closure decisions as visibility changes. Visual Whole-Body Control selects base-velocity and end-effector references from visual feedback for a lower-level controller \cite{vbc}. For aerial navigation~\cite{sketchprior}, Loquercio \emph{et al.} learn depth-based trajectory prediction from a privileged expert \cite{agileautonomy}. Our student also coordinates arm alignment, gripper timing, and post-acquisition stabilization using dual-view geometry and proprioceptive history throughout the task.

\section{Problem Statement}
\begin{figure}[!t]
\centering
\includegraphics[width=\columnwidth]{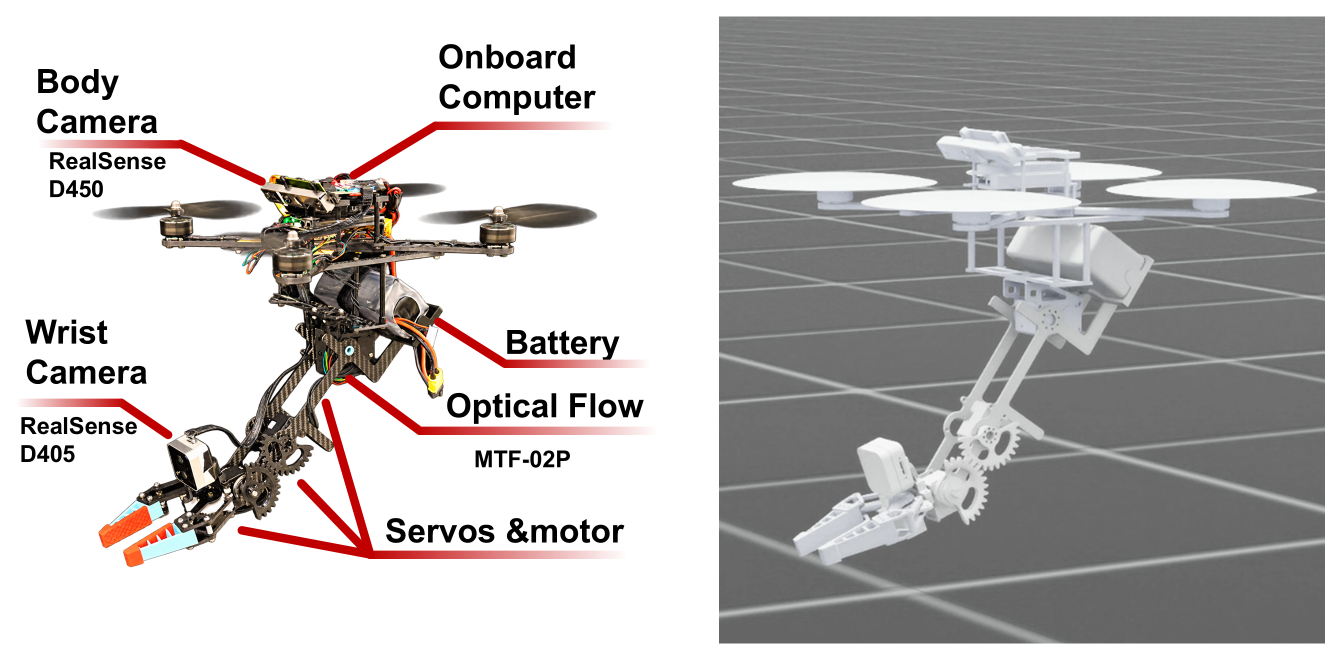}
\caption{Aerial manipulation platform. Left: physical prototype and camera placement. Right: simulation model used for training and evaluation.}
\label{fig:platform}
\vspace{-0.5cm}
\end{figure}

We build upon the QuadHand aerial manipulation platform~\cite{quadhand} and adapt it for simulated whole-body aerial acquisition and lifting. The system consists of an underactuated quadrotor, a three-DOF arm, and a two-finger gripper. Figure~\ref{fig:platform} shows the modified physical prototype, sensor placement, and simulation model.

Here, \emph{acquisition} denotes a simulated latch triggered by geometric, motion, posture, closure-command, and dwell conditions. We use this model because thin-finger contact simulation can be unstable and costly in large-scale parallel reinforcement learning. It retains the coupling between alignment, relative motion, closure timing, and payload loading. We formulate the task as a partially observable Markov decision process (POMDP), with observations \(\mathbf o_t\) and actions \(\mathbf a_t\). A single recurrent policy coordinates the aerial base, arm, and gripper through approach, acquisition, and lifting (Fig.~\ref{fig:task_sequence}), under environment-defined acquisition rules. Episodes start from \(\rho_0\) and end at the time limit or a configured failure. Success requires both the object and aerial base to rise by at least 0.15 m from their respective acquisition heights and satisfy the lift-and-hold conditions in Section V.

\begin{figure*}[!t]
\centering
\includegraphics[width=\textwidth]{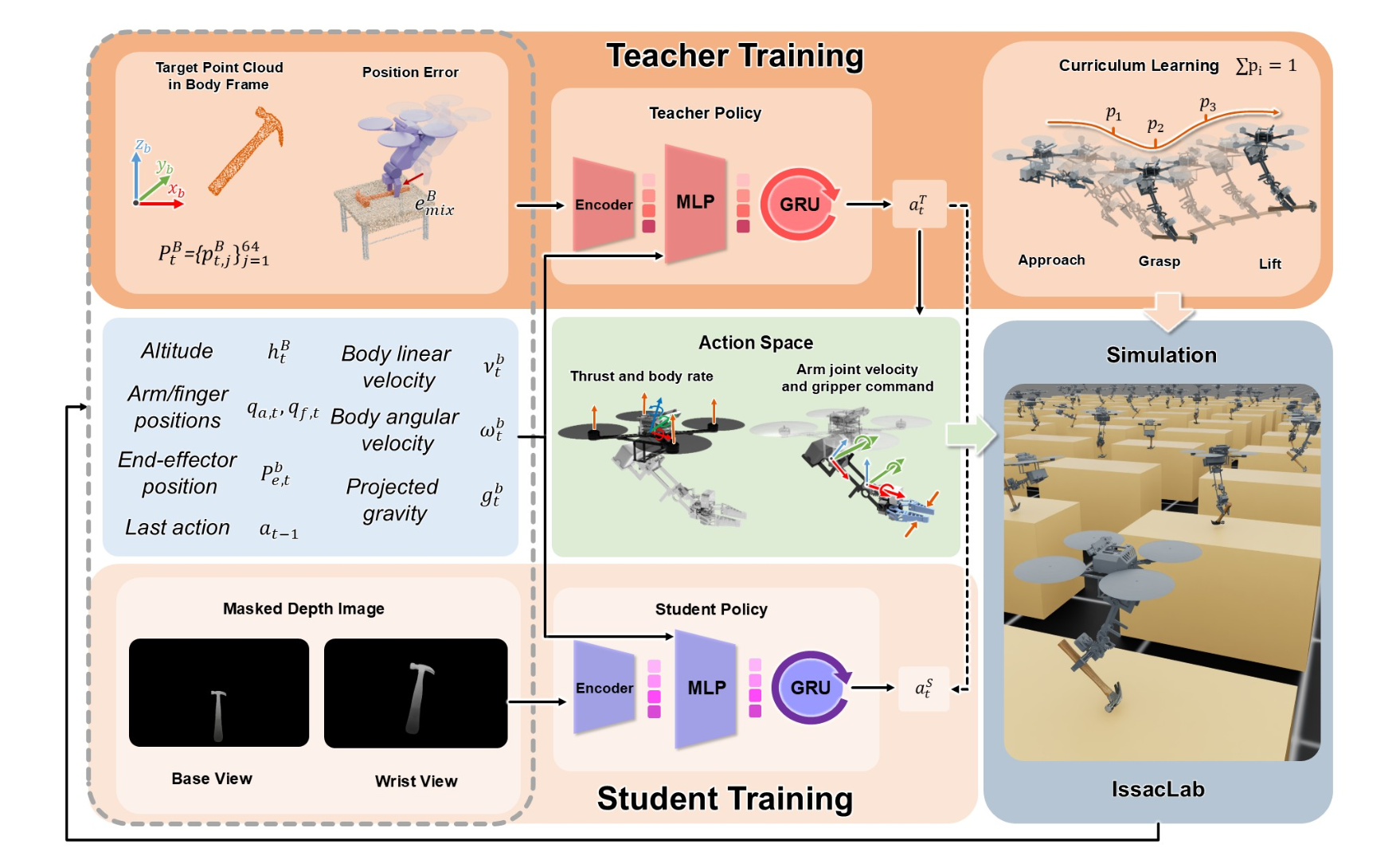}
\caption{Overview of the recurrent teacher--student framework. The privileged teacher is distilled into a recurrent visual student conditioned on dual-view point clouds and proprioception.}
\label{fig:method}
\end{figure*}

\section{Method}
We first train a privileged recurrent teacher with simulated target geometry, then transfer it to a student using dual-view camera observations (Fig.~\ref{fig:method}). Both policies use the same flight--arm--gripper action interface throughout the task.

\subsection{Reinforcement Learning from Target Geometry}

\subsubsection{Observation and action spaces}
At control step $t$, the teacher receives 64 points sampled from the target object and expressed in the UAM body frame $\mathcal B$, denoted by $\mathbf P_t^B\in\mathbb R^{64\times3}$. Its state vector $\mathbf z_t^{\mathrm T}\in\mathbb R^{33}$ contains body-frame linear and angular velocities, projected gravity, altitude, arm-joint and finger positions, end-effector position, and the previous action. The privileged components are the integrated base-position error, the blended position error $\mathbf e_{\mathrm{mix}}^B$ defined below, and the remaining object-height error. Superscripts $\mathrm T$ and $\mathrm S$ denote teacher and student quantities. Together, these inputs form a 225-dimensional observation. The eight-dimensional action is
\begin{equation}
\mathbf a_t=[a_T,a_{\omega x},a_{\omega y},a_{\omega z},
a_{q1},a_{q2},a_{q3},a_g]_t^{\mathsf T}.
\end{equation}
Collective thrust $a_T$ and body-rate commands $a_{\omega x},a_{\omega y},a_{\omega z}$ are executed by the low-level controller; arm commands $a_{q1},a_{q2},a_{q3}$ are integrated into joint-position targets, and $a_g$ commands the gripper.

\subsubsection{Reward design}
We train the teacher with proximal policy optimization (PPO) using approach, acquisition, and lifting rewards, together with visibility, motion regularization, and progress shaping.

\emph{Approach and braking:} With world frame $\mathcal W$, forward--left--up body frame $\mathcal B$, and arm joint-angle vector $\mathbf q_a\in\mathbb R^3$, end-effector kinematics and alignment error are
\begin{equation}
\begin{aligned}
\mathbf p_E^W&=\mathbf p_B^W+\mathbf R_{WB}\mathbf f_E(\mathbf q_a),\\
\mathbf e_E^B&=\mathbf R_{WB}^{\mathsf T}(\mathbf p_G^W-\mathbf p_E^W),
\end{aligned}
\end{equation}
where $\mathbf R_{WB}$ maps body-frame to world-frame coordinates. Forward kinematics $\mathbf f_E(\mathbf q_a)\in\mathbb R^3$ gives the end-effector position relative to the base origin in $\mathcal B$. Spatial superscripts identify frames; subscripts $B$, $E$, and $G$ denote the base, end effector, and grasp target. The body-frame error $\mathbf e_B^B$ measures displacement to the desired base location set by the grasp target and nominal grasp-arm configuration. For $d_E=\|\mathbf e_E^B\|_2$, the error $\mathbf e_{\mathrm{mix}}^B$ blends base positioning and end-effector alignment:
\begin{equation}
\begin{aligned}
\xi&=\operatorname{clip}\left(\frac{d_E-d_n}{d_f-d_n},0,1\right),\\
w&=1-3\xi^2+2\xi^3,\\
\mathbf e_{\mathrm{mix}}^B&=(1-w)\mathbf e_B^B+w\mathbf e_E^B .
\end{aligned}
\end{equation}
The near and far distances $d_n<d_f$ bound the transition; $\xi\in[0,1]$ is the normalized transition distance, and $w\in[0,1]$ is a smooth cubic interpolation weight increasing toward one near the target. For $d=\|\mathbf e_{\mathrm{mix}}^B\|_2$, a distance-dependent reference velocity provides braking guidance:
\begin{equation}
\begin{aligned}
v_{\max}(d)&=v_n+(v_f-v_n)(1-e^{-(d/d_s)^2}),\\
\mathbf v^\star&=\frac{\mathbf e_{\mathrm{mix}}^B}{\max(\|\mathbf e_{\mathrm{mix}}^B\|,\epsilon)}
\min\{k_p\|\mathbf e_{\mathrm{mix}}^B\|,v_{\max}(\|\mathbf e_{\mathrm{mix}}^B\|)\},
\end{aligned}
\end{equation}
where $v_n,v_f$ are speed limits, $d_s$ is the transition scale, $k_p$ is a position gain, and $\epsilon>0$ is a small distance in meters preventing division by zero. The measured guidance velocity is $\mathbf v_g=(1-w)\mathbf v_B^B+w\mathbf v_E^B$, where $\mathbf v_B^B$ and $\mathbf v_E^B$ are base and end-effector linear velocities relative to $\mathcal W$, expressed in $\mathcal B$. End-effector velocity includes base translation, rotation, and arm motion. For simulated latch state $L_t\in\{0,1\}$, the approach cost is
\begin{equation}
\begin{aligned}
r_{\mathrm{app}}={}&-\mathbb I[\neg L_t]\Delta t
\left(c_p\tanh\|\mathbf e_{\mathrm{mix}}^B/\mathbf s_p\|_2+
c_v\frac{z_v}{1+z_v}\right),\\
z_v={}&\|(\mathbf v_g-\mathbf v^\star)/s_v\|_2^2 .
\end{aligned}
\end{equation}
Here $\mathbb I[\cdot]$ is the indicator, $\Delta t$ is the control period, $\mathbf s_p,s_v$ normalize errors, and $c_p,c_v\geq0$ are weights; vector division is componentwise. Additional costs penalize excessive radial approach speed, near-target motion, and overshoot.

The configuration uses $d_n=0.075$ m, $d_f=0.45$ m, $v_n=0.08$ m/s, $v_f=0.5$ m/s, $d_s=0.2$ m, and $k_p=1$ s$^{-1}$. Position scales are $(0.2,0.1,0.1)$ m, the velocity scale is $0.2$ m/s, and the weights are $c_p=12$ and $c_v=40$.

\emph{Acquisition and lifting:} Readiness and closure-dwell rewards accompany a penalty for premature closure. Acquisition requires admissible geometry, precision, posture, clearance, altitude, and recent visibility, together with a sufficiently strong policy close command for three steps. Relative end-effector speed, body angular speed, and object motion must satisfy their limits within an eight-step window including the current step. After acquisition, lift-progress, height, uprightness, and holding terms reward the lift-and-hold endpoint defined in Section V.

\emph{Visibility, regularization, and progress:} Visibility shaping rewards target observability; motion and action-change costs regularize execution. Selected terms reward reductions in bounded error potentials or improvements over the episode's best progress. For designated terms, terminal failure cancels accumulated shaping credit, including the current step.

\subsubsection{Policy representation and termination}
A PointNet encoder maps the points in $\mathbf P_t^B$ to a 128-dimensional order-invariant feature. Concatenating this feature with $\mathbf z_t^{\mathrm T}$ feeds a $256$--$128$ MLP, a 128-unit GRU, and PPO action and value heads. The recurrent state resets each episode. The actor receives no task-phase label or acquisition flag.

Episodes end at the time limit or configured failures, including crashes, numerical failures, and drops when enabled. Time limits are value-estimation truncations. Controlled-approach success can terminate early curriculum episodes; acquisition does not terminate full-task rollouts.

\subsubsection{Critical-state curriculum}
We use four reset distributions: approach starts $\rho_0$, near-acquisition starts $\rho_n$, bridge starts $\rho_b$ linking approach to closure, and acquired-object starts $\rho_l$ for lifting. Their effective reset distribution at curriculum stage $k$ is
\begin{equation}
\begin{aligned}
\rho_k &= \alpha_{0,k}\rho_0+\alpha_{n,k}\rho_n
+\alpha_{b,k}\rho_b+\alpha_{l,k}\rho_l,\\
&\qquad \sum_{j\in\{0,n,b,l\}}\alpha_{j,k}=1 .
\end{aligned}
\end{equation}
The nonnegative $\alpha_{j,k}$ are effective sampling probabilities. Each reset has consistent base, arm, gripper, and object states; $\rho_l$ starts below the required lift height. Training progresses from controlled approach to closure and lifting with assisted resets, then extends bridge starts toward normal approach conditions. The final stage removes assisted resets and trains exclusively from $\rho_0$.

\subsection{Policy Transfer through Supervised Learning}

\subsubsection{Dual-view geometry and recurrent policy}
The student uses simulated Intel RealSense depth sensors: a body-mounted D450 module and a wrist-mounted D405 camera. The simulator supplies target instance masks; segmentation is outside the learned policy. We back-project masked depth pixels and transform the points into the body frame:
\begin{equation}
\widetilde{\mathbf p}^{B}=\mathbf T_{BC_i}(\mathbf q_a)
\begin{bmatrix}\Pi^{-1}(u,v,D_i;\mathbf K_i)\\1\end{bmatrix},
\end{equation}
where $i\in\{\mathrm{body},\mathrm{wrist}\}$ indexes the camera, $(u,v)$ is a target pixel, $D_i$ is depth, $\mathbf K_i$ contains intrinsics, and $\Pi^{-1}$ denotes back-projection. The homogeneous transform $\mathbf T_{BC_i}$ depends on arm configuration for the wrist camera. Each view supplies 64 point slots and validity masks.

A shared PointNet MLP encodes both views, followed by componentwise max pooling over all valid points:
\begin{equation}
\mathbf f_t=\max_{i,j:\,m_{t,i,j}=1}\phi(\mathbf p^B_{t,i,j}).
\end{equation}
Here $\mathbf f_t\in\mathbb R^{128}$ is the pooled feature, $j$ indexes point slots, $m_{t,i,j}$ indicates validity, and $\phi$ is the shared encoder. The pooled feature is zero if both views are invalid. The student state vector $\mathbf z_t^{\mathrm S}\in\mathbb R^{36}$ combines the 26 shared proprioceptive and previous-action components with two velocity-estimate quality indicators and four quality indicators per camera. The teacher's seven privileged components are omitted. An MLP $\psi$ fuses the inputs for a GRU with recurrent hidden state $\mathbf h_t\in\mathbb R^{128}$:
\begin{equation}
\mathbf h_t=\operatorname{GRU}
\left(\psi([\mathbf f_t,\mathbf z_t^{\mathrm S}]),\mathbf h_{t-1}\right).
\end{equation}
The point encoder has widths $3$--$64$--$128$--$128$, the fusion MLP has widths $256$--$128$, and the GRU has 128 hidden units. Two heads produce seven $\tanh$ flight--arm outputs and a gripper logit $\ell_t$, with $\ell_t>0$ requesting closure. An auxiliary head predicts target error during training; evaluation does not use it as a closure gate.

\subsubsection{Teacher initialization and closure supervision}
We initialize the student's point encoder, fusion MLP, GRU, and flight--arm head from compatible teacher parameters, dropping privileged input columns and zero-initializing new columns. Normalization retains the teacher statistics for shared inputs. The gripper head is initialized from the teacher's gripper output with reversed sign and a threshold shift, then trained with binary readiness labels.

Let $b_t\in\{0,1\}$ indicate that the instantaneous geometry, precision, relative-speed, body-rate, object-motion, posture, clearance, and altitude checks all pass. This model-defined readiness excludes the analytic visibility proxy and is distinct from the acquisition gate's recent-dynamics window and closure dwell. The closure supervision label $y_t\in\{0,1\}$ is
\begin{equation}
y_t=L_t\ \lor\ \bigwedge_{j=1}^{H}b_{t+j},\qquad H=3 .
\end{equation}
Thus closure is labeled positive for an acquired object or readiness throughout the next three steps, with the first five steps overridden to open. Future states are used for training labels; execution uses current and past observations.

For flight--arm action $\mathbf a_t^{fa}\in\mathbb R^7$ excluding the gripper, the objectives are
\begin{equation}
\begin{aligned}
\mathcal L_a&=\left\langle
\|\mathbf a_t^{fa}-\mathbf a_t^{fa,\mathrm{ref}}\|_W^2
\right\rangle,\\
\mathcal L_{\Delta a}&=\left\langle
\|\Delta\mathbf a_t^{fa}-\Delta\mathbf a_t^{fa,\mathrm{ref}}\|_2^2
\right\rangle ,
\end{aligned}
\end{equation}
where superscript $\mathrm{ref}$ denotes the teacher reference, $W$ is a positive diagonal weight matrix, $\|\mathbf x\|_W^2=\mathbf x^{\mathsf T}W\mathbf x$, and $\Delta\mathbf a_t^{fa}=\mathbf a_t^{fa}-\mathbf a_{t-1}^{fa}$. The teacher reference uses the same temporal difference. Brackets average over valid, sample-weighted time steps and sum channel contributions. Temporal differences require two adjacent valid samples.

The total loss is
\begin{equation}
\mathcal L=\lambda_a\mathcal L_a+
\lambda_\Delta\mathcal L_{\Delta a}+
\lambda_g\mathcal L_g+\lambda_e\mathcal L_e .
\end{equation}
The nonnegative $\lambda$ coefficients weight action matching, action-change matching, gripper classification, and target-error regression. $\mathcal L_g$ is positive-class-weighted binary cross-entropy on $\ell_t,y_t$; $\mathcal L_e$ is scaled squared target-error regression over valid pre-acquisition samples, with a distance cutoff in later stages. Sample weights emphasize near- and post-acquisition states, and sequence masks exclude padding.

For final consolidation, $W=\operatorname{diag}(4,1,1,1,2,2,2)$ and $(\lambda_a,\lambda_\Delta,\lambda_g,\lambda_e)=(1,0.2,0.5,1)$. The gripper positive-class weight is three. Target-error residuals in meters are multiplied by 20 before squaring, with supervision restricted to distances below 0.3 m.

\subsubsection{Behavioral cloning, DAgger, and consolidation}
We first use offline behavioral cloning on teacher rollouts, then collect learner-visited trajectories with DAgger-style aggregation. The teacher labels flight--arm actions at visited states; closure uses the look-ahead labels above. We alternate collection and supervised updates, then consolidate on accumulated data, including camera-randomized trajectories.

The archived student starts with 30 behavioral-cloning epochs at a learning rate of $10^{-4}$. Aggregation rounds use 12 supervised epochs at $5\times10^{-5}$; final consolidation uses 40 epochs at the same rate. Collection selects teacher actions with probability 0.25, 0.15, and 0.10 in the first three rounds and 0.05 thereafter. Evaluation uses only the frozen student, without teacher intervention.

\section{Experiments}

We evaluate the proposed framework through four aspects:
(1) curriculum learning for privileged policy acquisition,
(2) privileged-to-visual policy transfer,
(3) whole-body coordination during acquisition and lifting,
and (4) robustness under sensing, dynamics, payload, and object variations.
All experiments use the acquisition-and-payload model described below unless otherwise specified. We separately analyze pre-acquisition and post-acquisition lift-and-hold failures to distinguish acquisition from completion of loaded flight.

\subsection{Experimental Setup}

Training and primary evaluation use Isaac Lab; separate MuJoCo trials assess sim-to-sim transfer to additional object configurations. Main evaluations use a 10-ms physics timestep, 50-Hz policy control, and a nominal cylinder ($r=20$~mm, $h=95$~mm, $m=100$~g).

The physics/control domain randomization (DR) perturbs thrust dynamics, actuator delays, action latency, thrust noise, and inertial sensing noise and bias (Table~\ref{tab:dr}).

\begin{table}[!t]
\centering
\caption{Domain randomization settings.}
\label{tab:dr}
\footnotesize
\setlength{\tabcolsep}{4pt}
\renewcommand{\arraystretch}{1.0}
\begin{tabular}{@{}>{\raggedright\arraybackslash}p{0.46\columnwidth}
>{\raggedright\arraybackslash}p{\dimexpr0.54\columnwidth-2\tabcolsep\relax}@{}}
\toprule
Parameter & Setting\\
\midrule
Thrust-to-weight ratio & $[2.0,3.2]$\\
Thrust delay coefficient $\alpha$ & $[0.22,0.50]$\\
Body-rate delay coefficient $\alpha$ & $[0.33,0.63]$\\
Action delay & 1--3 control steps\\
Thrust noise std. & 5\%\\
IMU noise & Linear/angular velocity, gravity\\
IMU bias & Per-episode constant\\
\bottomrule
\end{tabular}
\end{table}

The main experiments use a configured acquisition-and-payload model. Acquisition requires admissible relative geometry, motion and posture, a policy-issued close command, and a three-step dwell. The nominal axial, lateral, and vertical acquisition tolerances are 25, 12, and 25 mm. After acquisition, the cup follows the gripper-link attachment frame, and a payload-reaction wrench models the gravitational and inertial load on the carrier. Finger--cup collision responses are filtered, so post-acquisition retention is imposed by the attachment model. These experiments do not validate frictional holding, force closure, or resistance to slip. Closure is irreversible, with a close-logit threshold of zero and the predicted-distance closure gate disabled.

Full-task success under this model requires acquisition followed by the configured lift-and-hold endpoint without a crash or drop. The endpoint includes at least 0.15 m of elevation of both the object and carrier relative to acquisition, the configured carrier and object height conditions, a body tilt of at most 8$^\circ$, a horizontal error of at most 0.10 m, and horizontal and absolute vertical speeds of at most 0.15 m/s. The completion conditions must persist for ten control steps (0.20 s). Acquisition, evaluator-classified crashes, alignment error, and relative speed provide complementary measurements of the task stages.

We evaluate under nominal conditions, physics/control DR, and physics/control DR with additional camera perturbations. Each main test uses 64 parallel environments for 2,000 control steps per rollout seed (1893, 2893, and 3893). Rates pool completed episodes after startup, including failures, and exclude episodes unfinished at rollout termination. Main and sensitivity tests evaluate the same frozen student checkpoint. Independent training seeds are used separately in the curriculum comparison.

\subsection{Comparison of Curriculum Learning Strategies}

We compare direct full-task training, a simple two-stage schedule, and the selected full-curriculum policy. Direct training starts from scratch on the full task; the two-stage baseline pretrains approach behavior before full-task training. In Table~\ref{tab:training}, each baseline uses three independent training seeds, with each final checkpoint evaluated using three rollout seeds under nominal and physics/control DR conditions. All three two-stage runs complete their 1,125-epoch budget.

\begin{table}[!t]
\centering
\caption{Full-task success under the acquisition-and-payload model for alternative training schedules. Direct and simple two-stage results are averaged across three independently trained policies; each policy is evaluated with three rollout seeds. The full-curriculum result is from the selected reference policy, evaluated with the same three rollout seeds.}
\label{tab:training}
\footnotesize
\begin{tabular}{lrr}
\toprule
Training schedule & Nominal (\%) & DR (\%)\\
\midrule
Direct full task & 33.33 & 32.59\\
Simple two stage & 0.00 & 0.00\\
Full curriculum & \textbf{100.00} & \textbf{99.90}\\
\bottomrule
\end{tabular}
\end{table}

\begin{figure}[!t]
\centering
\includegraphics[width=\columnwidth]{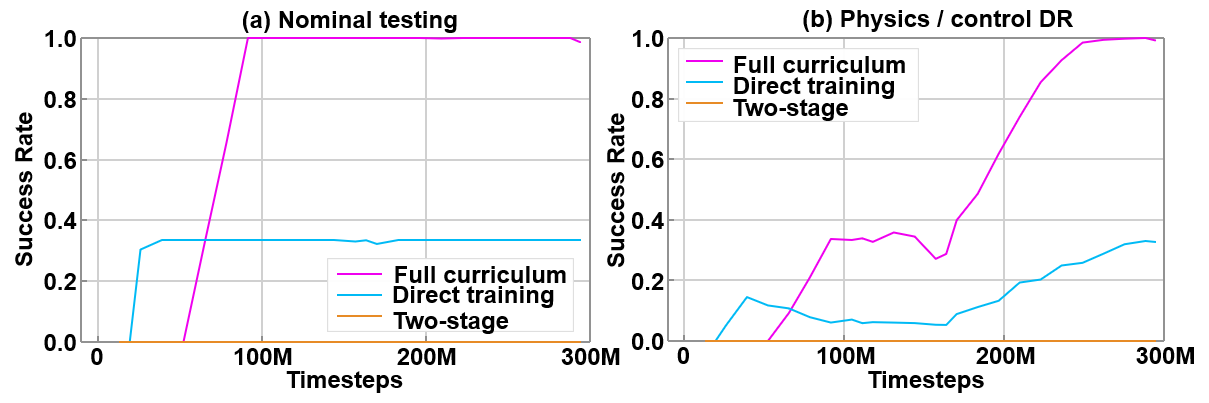}
\caption{Full-task success versus cumulative environment timesteps for three training schedules under (a) nominal and (b) physics/control domain-randomized conditions. Curves connect unsmoothed, equal-weight means across three training seeds; checkpoints missing any seed are omitted, breaking the lines.}
\label{fig:curriculum}
\end{figure}

Approach pretraining yields no full-task success in the tested two-stage runs; the selected full-curriculum reference exceeds the baseline averages (Table~\ref{tab:training}). This suggests that approach pretraining alone is insufficient under this schedule. Critical-state resets expose acquisition and lifting; bridge and normal-start training connect them to approach. Fig.~\ref{fig:curriculum} reports equal-weight means across three training seeds, with evaluation budgets varying by checkpoint.

\subsection{Visual Policy Learning and Teacher--Student Distillation}

We assess observation transfer by comparing the privileged Teacher, the BC-only visual Student after offline imitation, and the final Student after data-aggregation training.

\begin{table}[!t]
\centering
\caption{Full-task success under the configured acquisition-and-payload model. Each cell pools three evaluation seeds for one checkpoint; dashes denote unevaluated conditions.}
\label{tab:transfer}
\footnotesize
\setlength{\tabcolsep}{5pt}
\begin{tabular}{lrrr}
\toprule
Policy & Nominal (\%) & DR (\%) & + Camera (\%)\\
\midrule
Privileged teacher & 100.00 & 99.90 & ---\\
BC-only student & 0.00 & 0.00 & ---\\
Final student & 99.97 & 97.14 & 95.84\\
\bottomrule
\end{tabular}
\end{table}
Teacher denominators are 3,064 and 3,031 episodes; BC-only denominators are 7,105 and 8,271; final Student denominators are 3,057, 3,003, and 2,936, in table-column order.

The final Student trails the Teacher by 0.03 percentage points under nominal conditions and 2.76 points under physics/control DR (Table~\ref{tab:transfer}). Transfer retains near-teacher nominal performance without privileged target inputs, but the robustness gap widens under randomization.

The BC-only checkpoint yields no full-task successes after 30 offline imitation epochs. Learner-visited training supplies corrective supervision at states reached by the student. The final Student succeeds after this training, further optimization, and consolidation; the gain cannot be attributed to data aggregation alone.

Removing either camera sharply reduces success. Removing D405 leaves only the body-mounted D450 stream and yields no successes under nominal or combined physics/control-plus-camera-DR conditions. Using only the wrist-mounted D405 yields 12.40\% and 9.86\% success, respectively, well below the dual-view policy (Fig.~\ref{fig:sensitivity}(a)). Each condition uses three evaluation seeds. This tests the frozen policy's dependence on joint inputs; single-view variants are not retrained baselines.

\subsection{Alignment at Acquisition and Coordinated Execution}

At recorded latch events, we report latch-count-weighted means of per-seed end-effector alignment-error and relative-speed p90 values, where p90 denotes the 90th percentile. We also report the largest per-seed alignment-error p90. These weighted means are not percentiles of pooled samples.

The weighted alignment-error p90 is 8.12 mm under nominal conditions, 8.53 mm under physics/control DR, and 8.36 mm with additional camera DR. The corresponding worst-seed error p90 values are 8.16, 8.56, and 8.40 mm. Relative-speed p90 is 0.130, 0.138, and 0.138 m/s, respectively. These statistics are calculated from 3,056, 2,931, and 2,833 recorded latch samples in the same updated runs as Table~\ref{tab:transfer}.

Accepted-latch alignment changes little across profiles despite lower acquisition rates under randomization. Fewer episodes satisfy the acquisition conditions, but accepted events retain similar alignment. The gate excludes inadmissible states, so these statistics do not describe all approaches.

\begin{figure}[!t]
\centering
\includegraphics[width=\columnwidth]{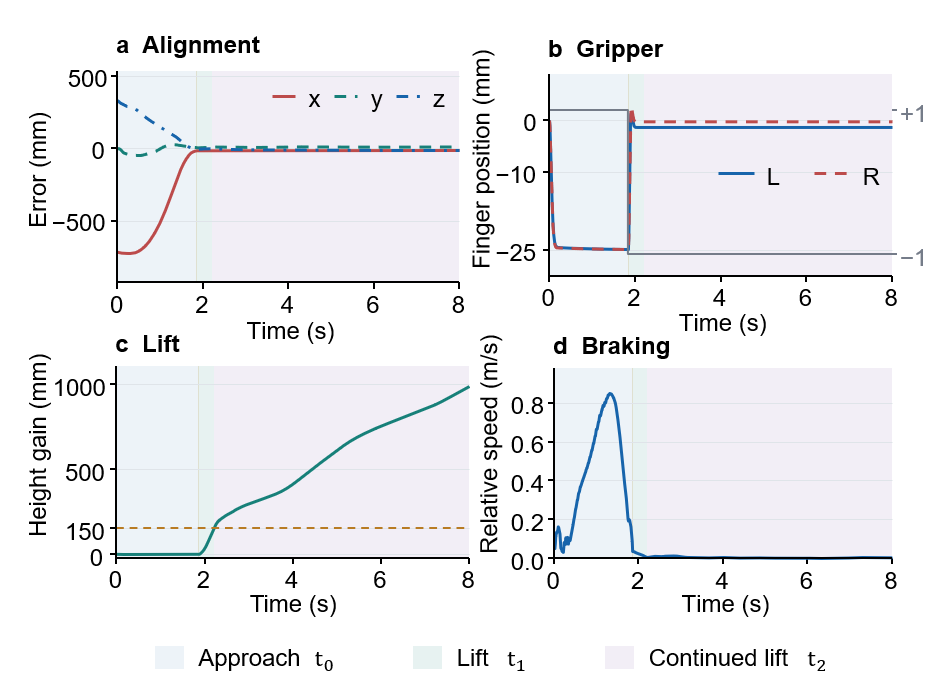}
\caption{Simulated acquisition and lifting sequence.
(a) End-effector alignment error.
(b) Finger positions and gripper command (right axis).
(c) Height gain; dashed line: 150-mm lift threshold.
(d) Relative end-effector speed.
Shading indicates execution phases: approach ($t_0$), acquisition ($t_1$), and lift ($t_2$), and is used for visualization only.}
\label{fig:execution}
\vspace{-0.5cm}
\end{figure}

Under nominal conditions, per-seed median times from episode start are 1.62--1.64 s for the close command, 1.66--1.68 s for aperture change, 1.70 s for acquisition, and 2.32 s for completion. These medians are not paired actuator-latency measurements. Figure~\ref{fig:execution} shows alignment error and relative speed decreasing before closure, followed by increasing finger positions and height gain. This matches the approach-and-braking objective and transition from closure to loaded lifting under one policy. Shaded phases describe execution and are not policy inputs.

A nominal diagnostic uses the same frozen Student in 16 environments for 1,000 control steps, yielding 121 completed episodes. Sampled points from both finger collision-mesh surfaces lie inside the cylindrical cup at the first latch frame in 121 episodes; bilateral overlap precedes latch in 63. This establishes geometric overlap, but filtered finger--cup collisions prevent conclusions about contact-force support, force closure, or stable physical retention.

\subsection{Robustness and Payload-Load Sensitivity}

\begin{figure}[!t]
\centering
\includegraphics[width=\columnwidth]{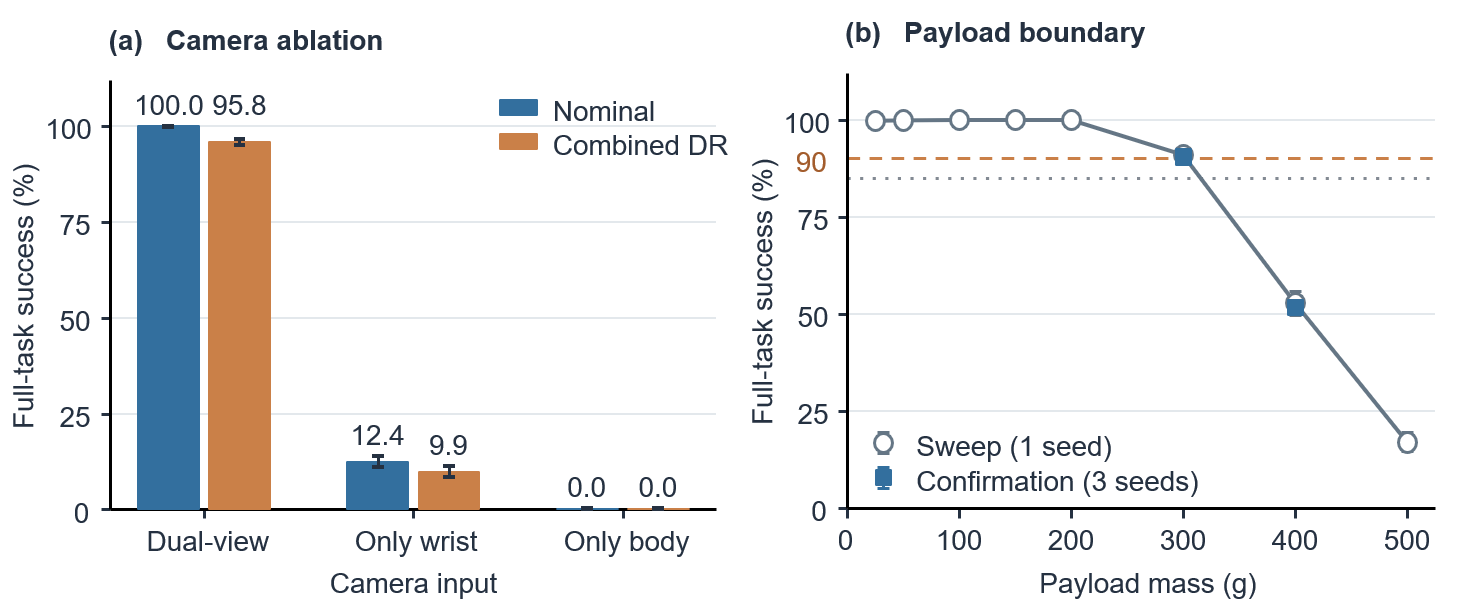}
\caption{Visual-input and payload sensitivity under the acquisition-and-payload model. (a) Camera removal from the frozen student without retraining; results pool three rollout seeds. Wrist-only uses D405; body-only uses D450. Combined DR includes physics/control and camera randomization. (b) Open circles: single-seed mass sweeps; filled squares: three-seed confirmations. Error bars: 95\% Wilson episode-level intervals. Dashed and dotted lines mark pooled and per-seed acceptance thresholds, respectively.}
\label{fig:sensitivity}
\end{figure}

\begin{figure}[!t]
\centering
\includegraphics[width=\columnwidth]{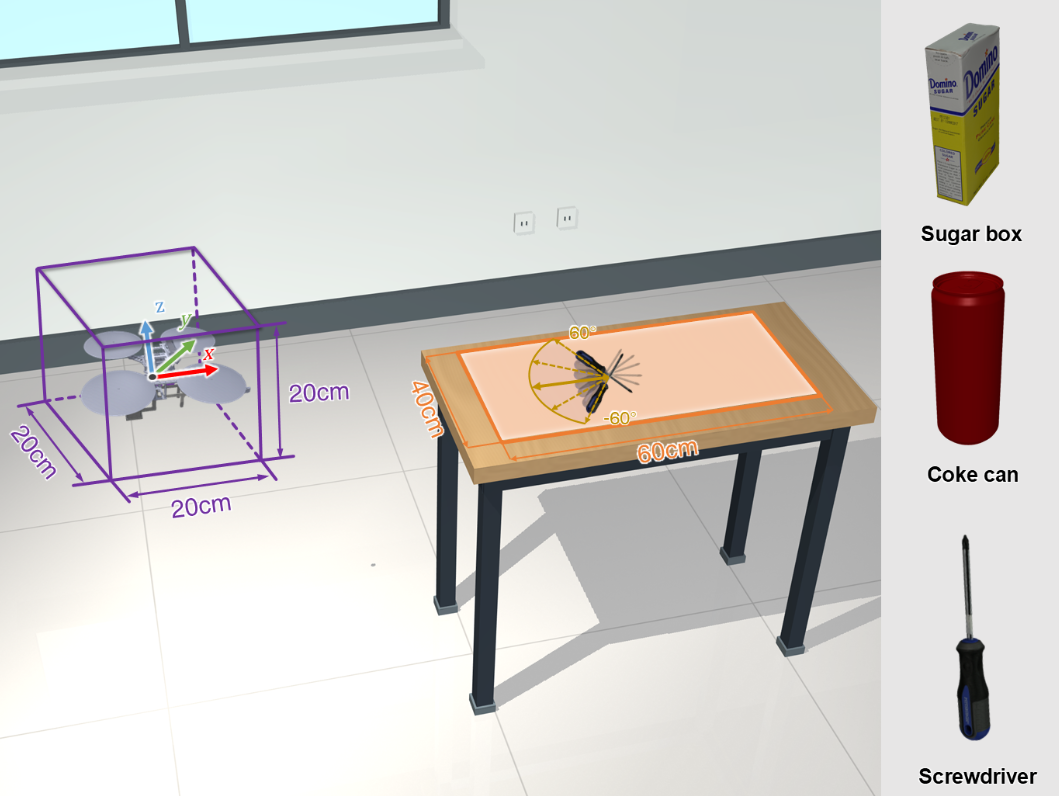}
\caption{MuJoCo evaluation setup. Left: candidate regions for randomized initialization of aircraft (purple) and object (orange) positions; sampled configurations are screened for initial target visibility. Right: the three test objects.}
\label{fig:mujoco_setup}
\end{figure}

\begin{figure*}[!t]
\centering
\includegraphics[width=0.93\textwidth]{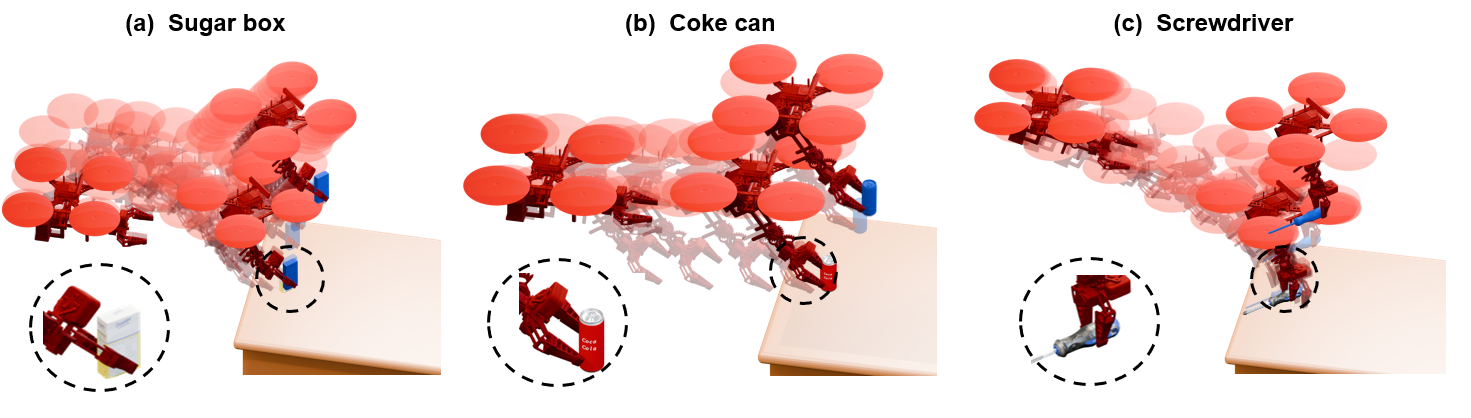}
\caption{Qualitative simulated acquisition-and-lift sequences for (a) a sugar box, (b) a Coke can, and (c) a screwdriver. Translucent overlays depict intermediate robot poses.}
\label{fig:objects}
\end{figure*}

For the nominal payload, the final Student maintains pooled success above 95\% under both randomized profiles (Table~\ref{tab:transfer}). Per-seed success ranges are 97.09--97.21\% under physics/control DR and 94.52--97.17\% with additional camera perturbations, indicating greater rollout variability in the latter condition.

Acquisition rates are 97.60\% under physics/control DR and 96.49\% with additional camera DR. Pre-acquisition crash rates are 2.40\% and 3.51\%, versus post-acquisition rates of 0.43\% and 0.51\%. These rates and the small acquisition-to-completion gap place most failures before acquisition under the nominal payload. After acquisition, the policy usually completes lift-and-hold.

We vary payload mass with fixed object geometry and policy weights to assess loaded task control (Fig.~\ref{fig:sensitivity}(b)). A coarse single-seed sweep tests 25, 50, 100, 150, 200, 300, 400, and 500 g. Tested masses from 25 to 200 g achieve 99.80--100\% success; rates fall to 91.03\%, 52.88\%, and 17.09\% at 300, 400, and 500 g, respectively. We confirm performance at 300 and 400 g using all three evaluation seeds. Each confirmation includes the sweep seed, so these results must not be pooled again.

Three-seed confirmation gives 90.50\% success at 300 g (2,754/3,043 episodes) and 51.75\% at 400 g (1,563/3,020 episodes). The respective per-seed ranges are 89.82--91.03\% and 49.30--53.07\%. The study-defined payload criterion requires at least 90\% pooled success and at least 85\% success for every evaluation seed.

The 300-g condition meets both criteria, whereas 400 g fails; 300 g is the largest tested passing three-seed confirmation point. This does not establish a continuous payload limit or success at untested masses.

Both confirmed payloads achieve 100\% acquisition without recorded crashes or drops. Lower full-task success at the heavier load thus reflects unmet post-acquisition lift-and-hold conditions, separating acquisition from loaded stabilization under the attachment-and-wrench model.

\subsection{Evaluation on Additional Objects}

Figure~\ref{fig:mujoco_setup} shows the MuJoCo setup. Initial aircraft and object positions are randomized within $20\times20\times20$ cm and $60\times40$ cm regions, respectively. Object yaw offsets are selected from $\{-60^\circ,-30^\circ,0^\circ,30^\circ,60^\circ\}$ relative to the nominal orientation.

Table~\ref{tab:object_generalization} summarizes 90 sim-to-sim trials across three objects; Fig.~\ref{fig:objects} illustrates their execution. The table note defines the success criteria.

\begin{table}[!t]
\centering
\caption{Acquisition-and-lift and full-task success in MuJoCo (30 trials per object).}
\label{tab:object_generalization}
\footnotesize
\setlength{\tabcolsep}{3pt}
\renewcommand{\arraystretch}{1.1}
\begin{tabular}{@{}>{\raggedright\arraybackslash}p{0.23\columnwidth}
>{\centering\arraybackslash}p{0.27\columnwidth}
>{\centering\arraybackslash}p{0.16\columnwidth}
>{\raggedright\arraybackslash}p{\dimexpr0.34\columnwidth-6\tabcolsep\relax}@{}}
\toprule
Object & \mbox{Acquisition--lift} & Full task & Failures\\
\midrule
Sugar box & 29/30 & 25/30 & 1 contact; 4 hold\\
Coke can & 28/30 & 28/30 & 2 contact\\
Screwdriver & 30/30 & 30/30 & None\\
\bottomrule
\end{tabular}
\par\vspace{2pt}
\begin{minipage}{\columnwidth}
\footnotesize
\emph{Note:} Objects share 30 presampled target positions. Acquisition--lift requires height gain $\geq10$ cm for $\geq0.5$ s, without unsafe states or robot--table contact. Full task requires height gain $(15\pm1.5)$ cm, object speed $<0.05$ m/s, and aircraft tilt $\leq15^\circ$ for $\geq2$ s; no unsafe states or robot--table contact are allowed throughout the trial. Contact denotes robot--table contact; hold denotes stable-hold failures. Sugar box and Coke can use virtual attachment; screwdriver uses native contact.
\vspace{-0.5cm}
\end{minipage}
\end{table}

Four sugar-box trials pass acquisition-and-lift but fail stable holding, so an initial-lift metric misses these failures. Both Coke-can failures involve robot--table contact; all trials passing acquisition-and-lift complete the full task. These outcomes distinguish holding failures from unsafe surface interaction. The screwdriver succeeds in all 30 native-contact trials. Different attachment mechanisms prevent attributing these differences to geometry alone.

\section{Conclusion and Future Works}
We presented a recurrent teacher--student policy that coordinates aerial acquisition and lifting without explicit task-phase inputs. The visual student maintains over 95\% pooled full-task success under tested simulation perturbations, with additional objects evaluated in MuJoCo. These results remain bounded by the stated interaction models; physical grasp retention is unvalidated. Future work will address camera calibration, online perception, sim-to-real transfer, and sustained transport under varied loads and dynamics.

\bibliography{ICRA2022}
\end{document}